\documentclass{article}

\usepackage{microtype}
\usepackage{graphicx}
\usepackage{booktabs} 
\usepackage{svg}
\usepackage{subcaption}
\usepackage{kantlipsum} 
\usepackage{amsmath,amssymb, stix}
\usepackage{todonotes}
\usepackage{wrapfig}
\usepackage{tikz}
\usetikzlibrary{matrix,positioning,quotes}
\definecolor{myblue}{RGB}{200,200,255}
\definecolor{mygreen}{RGB}{200,255,200}

\usepackage{hyperref}

\usepackage[accepted]{icml2021}

\icmltitlerunning{Graph-based neural acceleration for NMF}

\usepackage{amsmath,amsfonts,bm}

\def\eqref#1{equation~\ref{#1}}

\def\1{\bm{1}}

\def\ermH{{\textnormal{H}}}

\def\ermN{{\textnormal{N}}}

\def\ermV{{\textnormal{V}}}
\def\ermW{{\textnormal{W}}}

\def\ve{{\bm{e}}}

\def\vh{{\bm{h}}}

\def\vk{{\bm{k}}}

\def\vm{{\bm{m}}}

\def\vq{{\bm{q}}}

\def\vu{{\bm{u}}}
\def\vv{{\bm{v}}}
\def\vw{{\bm{w}}}
\def\vx{{\bm{x}}}

\def\mH{{\bm{H}}}

\def\mN{{\bm{N}}}

\def\mR{{\bm{R}}}

\def\mU{{\bm{U}}}
\def\mV{{\bm{V}}}
\def\mW{{\bm{W}}}

\def\mTheta{{\bm{\Theta}}}

\DeclareMathAlphabet{\mathsfit}{\encodingdefault}{\sfdefault}{m}{sl}
\SetMathAlphabet{\mathsfit}{bold}{\encodingdefault}{\sfdefault}{bx}{n}

\def\gE{{\mathcal{E}}}

\def\gG{{\mathcal{G}}}

\def\gL{{\mathcal{L}}}

\def\gV{{\mathcal{V}}}

\def\emV{{V}}

\begin{document}

\twocolumn[
\icmltitle{Graph-based Neural Acceleration\\for Nonnegative Matrix Factorization}



\icmlsetsymbol{equal}{*}

\begin{icmlauthorlist}
\icmlauthor{Jens Sjölund}{equal,UU}
\icmlauthor{Maria Bånkestad}{equal,UU,RISE}
\end{icmlauthorlist}

\icmlaffiliation{UU}{Department of Information Technology, Uppsala University, Uppsala, Sweden}
\icmlaffiliation{RISE}{Research Institutes of Sweden (RISE), Stockholm, Sweden}

\icmlcorrespondingauthor{Jens Sjölund}{jens.sjolund@it.uu.se}

\icmlkeywords{Graph Neural Network, Data-driven optimization, Nonnegative Matrix Factorization}

\vskip 0.3in
]



\printAffiliationsAndNotice{\icmlEqualContribution} 

\begin{abstract}
We describe a graph-based neural acceleration technique for nonnegative matrix factorization that builds upon a connection between matrices and bipartite graphs that is well-known in certain fields, e.g., sparse linear algebra, but has not yet been exploited to design graph neural networks for matrix computations. We first consider low-rank factorization more broadly and propose a graph representation of the problem suited for graph neural networks. Then, we focus on the task of nonnegative matrix factorization and propose a graph neural network that interleaves bipartite self-attention layers with updates based on the alternating direction method of multipliers. Our empirical evaluation on synthetic and two real-world datasets shows that we attain substantial acceleration, even though we only train in an unsupervised fashion on smaller synthetic instances. 

\end{abstract}

\section{Introduction}
Nonnegative matrix factorization (NMF) is a standard tool for analyzing high-dimensional nonnegative data such as images, text, and audio spectra \citep{Fu2019, Gillis2020}. In NMF, the aim is to find an approximate low-rank factorization of a given data matrix $\mV\in\mathbb{R}^{m\times n}$ according to
\begin{equation}
    \mV\approx \mW\mH^\top, \quad \mW\geq 0, \mH\geq 0,
\end{equation}
where the basis matrix $\mW\in\mathbb{R}^{m\times r}$ and the mixture matrix $\mH\in\mathbb{R}^{n\times r}$ are elementwise nonnegative matrices of rank $r\leq\min(m,n)$. 

The nonnegativity constraints are the source of both the interpretability and the complexity of NMF. In their absence, the optimal solution can be computed efficiently using e.g., a singular value decomposition (SVD). In their presence, the problem becomes NP-hard \citep{Vavasis2010}. State-of-the-art algorithms for general NMF are essentially direct applications of standard optimization techniques \citep{Huang2016,Gillis2020}, which are sensitive to the initialization and may require many iterations to converge.

Like other data-driven optimization methods \citep{Li2016,Chen2021}, we exploit that this problem has a specific structure, which we can \emph{learn} to take advantage of. However, a practically useful NMF method must be able to factorize matrices in a wide range of sizes, which means that---unlike most other applications where data-driven optimization has been used---our method must be able to solve problems of \emph{variable} size. 

To the best of our knowledge, such methods have previously only been explored for combinatorial optimization problems, which are often stated directly in terms of graphs and are therefore well-suited to graph neural networks \citep{Gasse2019,Cappart2021}. There is, however, a direct connection also between matrices and graphs. This connection is well-known in certain fields, e.g. sparse linear algebra \citep{Kepner2011, Duff2017}, but has not been exploited to design graph neural networks that learn to perform matrix computations---until now.

We describe a graph-based neural acceleration technique for nonnegative matrix factorization. Our main contributions are as follows, we (i) formulate low-rank matrix factorization in terms of a weighted bipartite graph; (ii) propose a graph neural network for alternating minimization, which interleaves Transformer-like updates with updates based on the alternating direction method of multipliers (ADMM); (iii) demonstrate substantial acceleration on synthetic matrices, also when applied to larger matrices than trained on; and (iv) modify the synthetic data generation to achieve successful generalization to two real-world datasets.

\section{Related Work}
Nonnegative matrix factorization has a wide range of applications, for instance in image processing \citep{Fu2019}, bioinformatics \citep{Kim2007}, and document analysis \citep{Shahnaz2006}, in part because it tends to produce interpretable results \citep{Lee1999}. Since the problem is NP-hard \citep{Vavasis2010}, 
this has led to the development of several algorithms specialized for NMF. A few of them exploit the special analytical structure of the problem \citep{Recht2012, Gillis2013}, but the majority are adaptations of standard optimization methods \citep{Lin2007,Kim2008,Xu2013,Gillis2020}.

In contrast, learning to optimize is an emerging approach where the aim is to learn optimization methods specialized to perform well on a set of training problems \citep{Chen2021}. The same approach is also referred to as data-driven optimization in more mathematically oriented works \citep{Banert2020}, or as meta-learning or learning-to-learn when referring to the more narrow scope of learning parameters for machine learning models \citep{Li2016,Andrychowicz2016,Bello2017}. It is related to automated machine learning, ``AutoML'' \citep{Hutter2019}, but rather than only selecting from or tuning a few existing optimizers, learning to optimize generally involves a search over a much larger space of parameterized methods that encompass entirely new optimization methods \citep{Maheswaranathan2020}.

The iterative nature of optimization methods is often captured using either feedforward or recurrent neural networks. In the feedforward approach, which we use, the algorithm is unrolled over a limited number of iterations and intertwined with learned components \citep{Gregor2010,Domke2012,Monga2021}. This makes it easier to train than the recurrent approach \citep{Andrychowicz2016,Venkataraman2021}, but computational considerations limit the number of iterations (depth) that can be considered.

Data-driven optimization methods applied to image-based problems \citep{Diamond2017,Banert2020} typically rely on convolutional neural networks (CNNs). However, the geometry of matrices, when used in computations, is decidedly different from images. Matrices have a non-Euclidean structure and violate the assumption of translational equivariance that underlies regular CNNs. This motivates the need to use tools from geometric deep learning \citep{Bronstein2017}, and graph convolutional networks in particular \citep{Kipf2016,Gilmer2017}.

For similar reasons, there has been a recent surge of interest in graph neural networks for combinatorial optimization problems \citep{Dai2017,Gasse2019,Nair2020,Cappart2021}, which are then often represented as bipartite graphs between variables and constraints. Another problem that is naturally represented as a bipartite graph is (low-rank) matrix completion, which has, indeed, also been successfully tackled with graph neural networks \citep{vanDenBerg2017, Monti2017}.

There is a vast literature building upon the connection between graphs and matrices, both as an efficient way to implement graph algorithms \citep{Kepner2011} and as a theoretical tool for (sparse) linear algebra \citep{Brualdi2008,Duff2017}. However, to the best of our knowledge, no one has previously used it to design graph neural networks for numerical linear algebra.

\section{Background}
In the literature on nonnegative matrix factorization, the most common way to measure the quality of the approximation, $V\approx WH^\top $, is the Frobenius norm, 
leading to the optimization problem 
\begin{equation}
    \begin{aligned}
        & \underset{\mW, \mH}{\text{minimize}}
        & & \frac{1}{2}\|\mW\mH^\top   - \mV\|_\text{F}^2 \\
        & \text{subject to}
        & & \mW, \mH \geq 0.
    \end{aligned}\label{eq:standardNMF}
\end{equation}
If either of the factors $\mW$ or $\mH$ is held fixed, what remains is a nonnegative least-squares problem, which is convex. 
For this reason, most NMF methods use an alternating optimization scheme \citep{Lee2001, Xu2013, Gillis2020}. As these methods are---at best---guaranteed to converge to a stationary point, it is important to initialize them appropriately. Popular initialization methods include nonnegative double SVD \citep{Boutsidis2008} and (structured) random methods \citep{Albright2006}. Our method provides a factorization that can be used by itself or as a refined initialization that other methods can process further. Its underpinning is a representation of matrix factorization as a bipartite graph. 

\subsection{Matrices as Graphs}
Consider a graph, $\gG = (\gV, \gE)$, with nodes $v_i\in\gV$, and directed edges $e_{i,j}\in\gE$ going from a source node $v_i$ to a target node $v_j$. An undirected graph can be modeled as having directed edges in both ways. A path is a sequence of distinct nodes connected by directed edges.
\begin{figure}[ht]
    \centering
    \includegraphics[width=0.8\columnwidth]{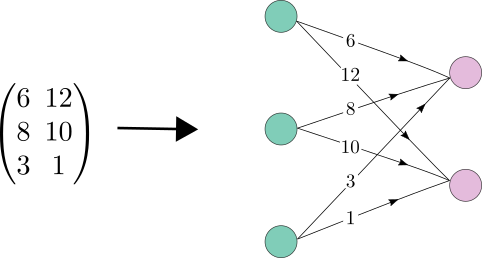}
    \caption{A matrix corresponds to a weighted bipartite graph.}
    \label{fig:matrix2graph}
\end{figure}
We may enrich the graph by associating features, $\vx_i\in\mathbb{R}^p$ and $\ve_{i,j}\in\mathbb{R}^q$, to the nodes and edges, respectively. 

The starting point for our method is the fact that any matrix $\mV\in\mathbb{R}^{m\times n}$ can be represented as a directed, weighted, and bipartite graph $\gG(\mV)$ called the König digraph \citep{Doob1984}. The König digraph has $(m+n)$ nodes---$m$ row nodes and $n$ column nodes---with edges going from row nodes to column nodes.  The weight of each edge is given by the corresponding element in the matrix, $\ve_{i,j}=\emV_{i,j}$, i.e. the edge features are scalar-valued ($q=1$). 
An example is shown in Figure \ref{fig:matrix2graph}. By convention, edges corresponding to zeros in the matrix are omitted.
Transposing a matrix corresponds to reversing the edges in the associated König digraph. The weight of a path is the product of the weights along the path.

\subsection{Message Passing Neural Networks}
The recent surge of interest in graph neural networks has mainly focused on message passing neural networks \citep{Gilmer2017} which encompasses graph convolutional networks \citep{Kipf2016}. In message passing neural networks, a forward pass consists of multiple iterations of message passing, optionally followed by a readout. Each message passing iteration $t$ consists of evaluating message functions $\vm^t_{i\to j}$ for every edge $\ve_{i,j}$, aggregating these at each node using a permutation-invariant aggregation function $\dottedsquare$ (typically the arithmetic mean or sum) and updating the node features using an update function $\vu^t$,
\begin{align}
    \vm^{t+1}_j &= \underset{i\in\mathcal{N}(j)}{\dottedsquare} \vm^t_{i\to j}\left(\vx_i^t, \vx_j^t, \ve_{i,j} \right),\\
    \vx^{t+1}_j &= \vu^t\left(\vx^t_j, \vm^{t+1}_j\right),
\end{align}
where $\mathcal{N}(j)$ denotes the neighbours of node $v_j$. 
In the readout phase, a readout function is used to compute a feature vector for the whole graph, but since we work directly with the node features in this work we do not use a readout. 

\section{Graph Representation of Constrained Low-rank Factorization}\label{sec:graph_representation}
We will now describe our graph representation of the NMF problem, which is generally applicable to constrained low-rank factorization. By representing the problem in a format suitable for graph convolutional neural networks, we make it possible to learn a fitting method tailored to the problem. In particular, our representation of the problem as a bipartite graph makes it possible to implement the graph neural network as a neural acceleration scheme for conventional alternating optimization methods. 
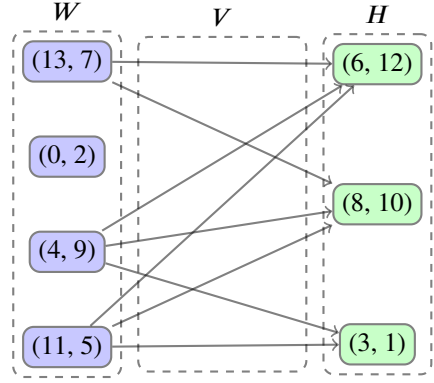
\begin{figure}
\centering
\begin{tikzpicture}[thick,amat/.style={matrix of nodes, 
  row sep=2em, draw, dashed, rounded corners, draw opacity=0.5,
  nodes={draw,solid, rectangle, rounded corners}},
  row_node/.style={fill=myblue}, 
  col_node/.style={fill=mygreen}]

 \matrix[amat,nodes=row_node, label=above:$W$] (W)
 {(13, 7)\\
  (0, 2)\\
  (4, 9)\\
  (11, 5)\\};
 
 \matrix[amat, right=2mm of W, label=above:$V$] (V) {
 &[1cm]&[1cm]\\[2em]
 &&\\[2em]
 &&\\[2em]
 &&\\[2em]
 };

 \matrix[amat,right=2mm of V,nodes=col_node,label=above:$H$] (H) {
 (6, 12)\\[1.7em] 
 (8, 10)\\[1.7em]
 (3, 1)\\};

 \draw[->, opacity=0.5]
    (W-1-1) edge (H-1-1)
    (W-1-1) edge (H-2-1)
    (W-3-1) edge (H-1-1)
    (W-3-1) edge (H-2-1)
    (W-3-1) edge (H-3-1)
    (W-4-1) edge (H-1-1)
    (W-4-1) edge (H-2-1)
    (W-4-1) edge (H-3-1);
\end{tikzpicture}
\caption{A simple example of how we represent a matrix factorization $\mV=\mW\mH^\top  $ as a bipartite graph. To reduce clutter, the edge weights $V_{ij}$ are not shown.} \label{fig:factorizedGraph}
\end{figure}
Before going into the derivation, consider the following factorization
\begin{equation}
\underbrace{
    \begin{pmatrix}
        162 & 174 & 46 \\
        24 & 20 & 2 \\
        132 & 122 & 21 \\
        126 & 138 &38
    \end{pmatrix}
    }_\mV
    =
    \underbrace{
    \begin{pmatrix}
        13 & 7 \\
        0 & 2 \\
        4 & 9 \\
        11 & 5
    \end{pmatrix}
    }_\mW
    \underbrace{
    \begin{pmatrix}
        6 & 8 & 3\\
        12 & 10 & 1
    \end{pmatrix}
    }_{\mH^\top  },
\end{equation}
that we represent as a bipartite graph according to Figure \ref{fig:factorizedGraph}.

\begin{figure*}
    \centering
    \includegraphics[width = \linewidth]{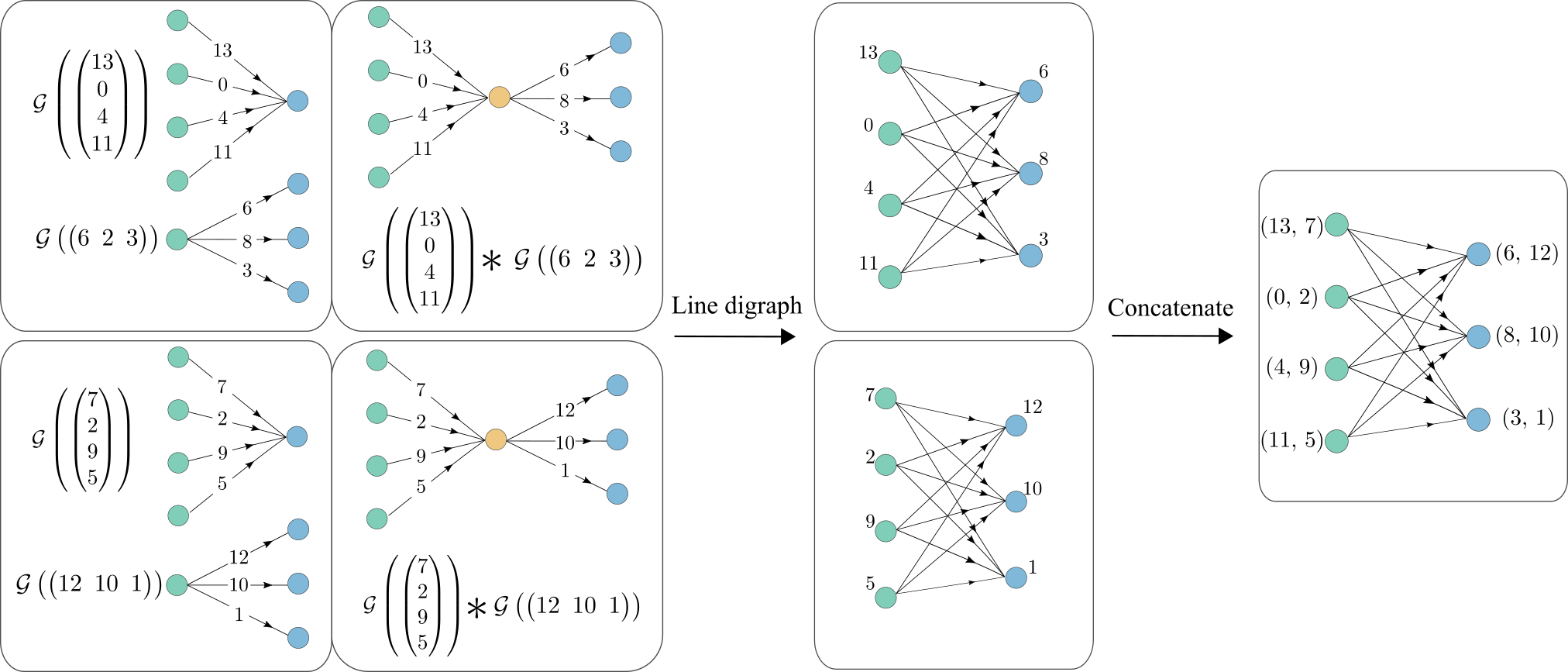}
    \caption{We represent matrix multiplication as a graph by expressing it as a sum of outer products, which are processed separately until the final, concatenation, step. The processing consists of (i) representing each vector as a König digraph, (ii) forming the concatenation graph, (iii) mapping each concatenation graph to the corresponding line digraph, and (iv) concatenating the node features of the line digraphs.}
    \label{fig:derivation_graph_representation}
\end{figure*}
Matrix multiplication can be expressed in terms of König digraphs. Given $\mW\in\mathbb{R}^{m\times r}$ and $\mH\in\mathbb{R}^{n\times r}$, we first define the concatenation graph $\gG(\mW)*\gG(\mH^\top  )$ by identifying the $r$ column nodes of $\gG(\mW)$ with the $r$ row nodes of $\gG(\mH^\top  )$. Visually, this corresponds to ``gluing together'' the digraphs. The $(i,j)$ entry of the matrix product $\mW\mH^\top  $ is the sum of the weights of all paths in the concatenation graph $\gG(\mW)*\gG(\mH^\top  )$ from row node $i$ to column node $j$. 

Our derivation, illustrated in Figure \ref{fig:derivation_graph_representation}, starts with expressing matrix multiplication as a sum of outer products,
\begin{equation}
    \mW\mH^\top = \sum_{c=1}^r \vw_c \vh_c^\top  ,
\end{equation}
where $\vw_c$ and $\vh_c$ denotes column $c$ in $\mW$ and $\mH$, respectively. 
The concatenation graph corresponding to an outer product has a unique path between each pair of row and column nodes. Hence, performing matrix multiplication by summing over outer products can be viewed as creating $r$ such concatenation graphs, computing path weights in each graph separately, and then summing across them.

To convert this into a format suitable for a graph neural network, we need to accomplish two additional things. First, we need to turn the paths into edges or nodes. Second, we need to encode the matrix $\mV$ that we wish to factorize.

The first objective can be achieved by converting each outer product graph $\gG\left(\vw_c\vh_c^\top  \right)$ to its corresponding line digraph $\gL\left(\gG\left(\vw_c\vh_c^\top  \right)\right)$ \citep{Harary1960}. This is again a directed bipartite graph with $m$ row nodes and $n$ column nodes, but the node features in each set correspond to the components of $\vw_c$ and $\vh_c$, respectively, and the edges correspond to the paths in $\gG\left(\vw_c\vh_c^\top  \right)$. 

A directed line graph is unweighted but, since it transforms paths into edges, it essentially transforms path weights into edge properties. An element in the outer product can be computed simply by multiplying the node features of the source and the target nodes. This suggests that if we concatenate the node features of the line digraphs $\gL\left(\gG\left(\vw_c\vh_c^\top  \right)\right)$ corresponding to each outer product $c=1,\ldots, r$, then an entry $(i,j)$ of the matrix product $\mW\mH^\top  $ can be computed on the edge $\ve_{i,j}$ of the resulting graph. Namely, by first computing the elementwise product of the node features and then summing the result. Further, if we define $\ve_{i,j}=\mV_{i,j}$, we can compare the current value of the product with the desired one, thus resolving the second issue of encoding the matrix $\mV$ as well. We will refer to the resulting graph as an augmented line digraph.


\section{Optimization Method}

There is an inherent symmetry between the factors in the NMF problem since the objective function (loss) $\ell(\mW, \mH; \mV)$ of the optimization problem in Equation \ref{eq:standardNMF} is invariant under transposition 
\begin{equation}
\ell(\mW, \mH; \mV) \triangleq \frac{1}{2}\|\mW\mH^\top   - \mV\|_F^2 = \ell(\mH, \mW; \mV^\top  ).
\end{equation}
Again, keeping either $\mW$ or $\mH$ fixed results in a nonnegative least-squares problem. Due to symmetry, we can use the same implementation for both subproblems, and we will therefore only describe the $\mH$-update,
\begin{equation}
    \hat{\mH} = \underset{\mH\geq 0}{\text{argmin}} \frac{1}{2}\|\mW\mH^\top   - \mV\|_\text{F}^2.\label{eq:subproblem}
\end{equation}
This can also be interpreted using our graph representation. The $\mH$-update corresponds to Figure \ref{fig:factorizedGraph}, where the information flows from $\mW$ to $\mH$. Conversely, the information flows in the opposite direction in the $\mW$-update, which uses $\mV^\top  $.


Following \citet{Huang2016}, we apply the alternating direction method of multipliers (ADMM) \citep{Boyd2011} to the subproblems, but only unroll a few iterations of it instead of solving the subproblems exactly at each iteration. See Appendix \ref{sec:ADMM} for details. Empirically, such early stopping has been found to produce better local minima \citep{Udell2016}. In addition to being faster, other major advantages of adopting ADMM with early stopping for neural acceleration, as compared to a generic convex optimization layer \citep{Agrawal2019}, is that it supports warm starting and problems of variable size. 

Various extensions to (i) losses other than the Frobenius norm, and (ii) constraints other than nonnegativity, are possible within essentially the same framework \citep{Huang2016, Udell2016}. We foresee that the neural acceleration method we outline in the next section would be relatively straightforward to extend in a similar fashion.

\section{Neural Acceleration}
\begin{figure}

\centering
\begin{subfigure}{.99\columnwidth}
\centering
\includegraphics[width=0.98\columnwidth]{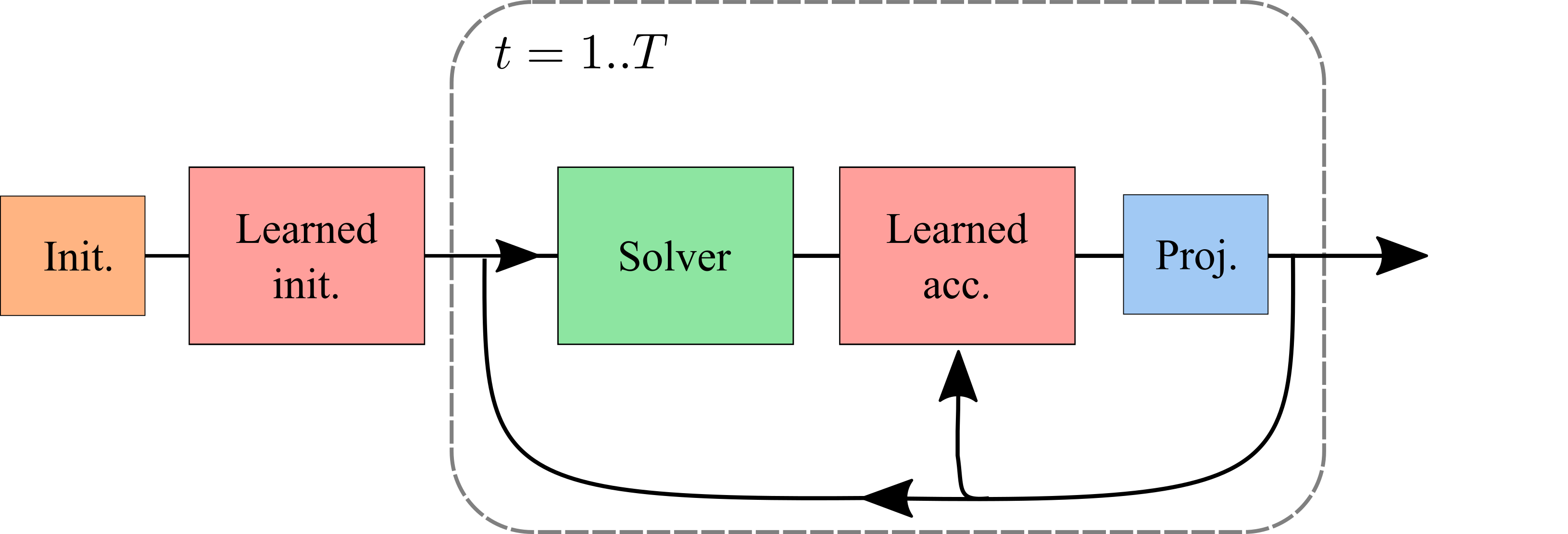}%
\caption{General model.}%
\label{fig:overview_general_acc_model}%
\end{subfigure}\hfill%
\vspace{0.2in}
\begin{subfigure}{.99\columnwidth}
\centering
\includegraphics[width=0.78\columnwidth]{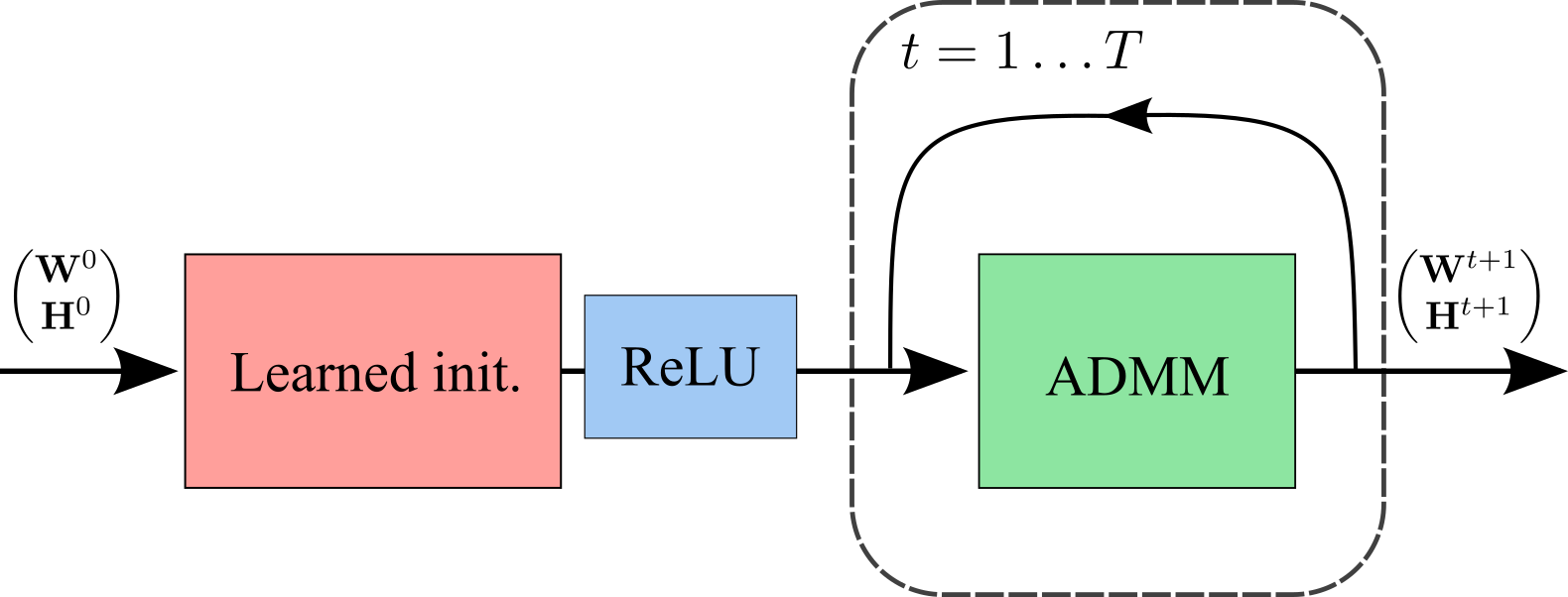}%
\caption{Learned initialization model.}%
\label{fig:overview_init_model}%
\end{subfigure}\hfill%
\vspace{0.2in}
\begin{subfigure}{.99\columnwidth}
\centering
\includegraphics[width=0.85\columnwidth]{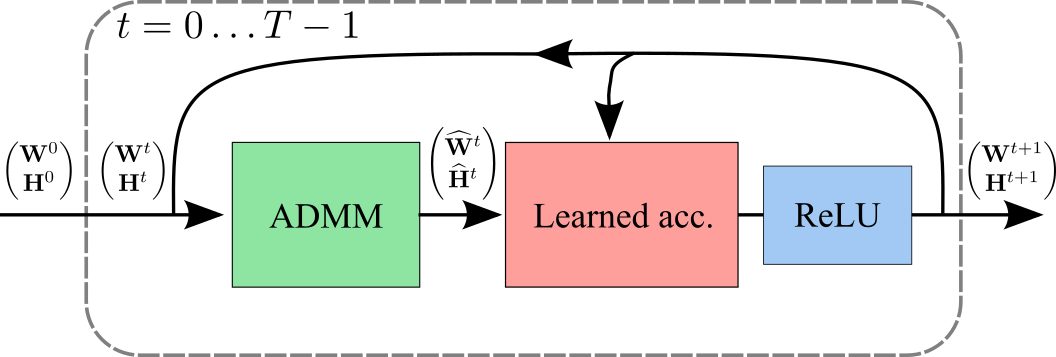}%
\caption{Learned acceleration model.}%
\label{fig:overview_acc_model}%
\end{subfigure}%
\caption{We discuss the general model for neural acceleration shown in (a), and perform experiments using the learned initilization model in (b) and the neural acceleration model in (c).}%
\label{fig:models}%
\end{figure}

We propose a neural acceleration method for NMF based on the bipartite graph representation in Section \ref{sec:graph_representation}. The overall structure of the model is shown in Figure \ref{fig:overview_general_acc_model}. The acceleration is achieved by a graph neural network that adapts the exceedingly successful Transformer architecture \citep{Vaswani2017} to weighted graphs.

Bipartite graphs appear in many application areas, so it is not surprising that Transformers has previously been extended to this setting \citep{Shu2020, hudson2021ganformer}. However, these works, as well as the original graph attention network \citep{Velickovic2017}, place little (if any) emphasis on the edge features. In our graph representation, the edge features are undoubtedly important, since they correspond to the data matrix that we wish to factorize. To place node and edge features on equal footing, we, therefore, extend the relation-aware self-attention mechanism of \citet{Shaw2018} to arbitrary directed, weighted, graphs. We also show how this can be combined with implicit edge features that are computed dynamically, in what we refer to as a factor Transformer layer or Factormer for short.

Multiple Factormers are then combined into an $N$-Factormer, which updates the factors in an alternating fashion using both Factormers and unrolled ADMM iterations. The constraints are enforced in the final layer by performing a Euclidean projection on the feasible set, which in the case of NMF simply amounts to setting negative entries to zero using a rectified linear unit. More generally, as long as the feasible set is convex, the Euclidean projection can be expressed as a convex quadratic program that could be incorporated as a layer in the network \citep{Agrawal2019}.

\subsection{Implicit Edge Features}
According to the graph representation in Section \ref{sec:graph_representation}, the edge features correspond to elements in the data matrix $\ve_{i,j}=\mV_{i,j}$, and are therefore static. However, during execution, we dynamically compute implicit edge features by augmenting the edge features with the element-wise products of the node features 
\begin{equation}
    \tilde{\ve}_{i,j}(\vx_i, \vx_j, \ve_{i,j}) = 
    \left[\vx_i \odot \vx_j, \ve_{i,j}\right],
    \label{eq:implicit_edge}
\end{equation}
where $\odot$ denotes element-wise product.
This is motivated by the fact that the residual $\mR_{i,j}$ associated with edge $(i,j)$ is given by 
\begin{equation}
    \begin{aligned}
        \mR_{i,j} &= \left(\sum_{l=1}^r \mW_{i,l}\mH_{j,l}\right)-\mV_{i,j}\\
                  &= \tilde{\ve}_{i,j}\left(\mW_{i, :}, \mH_{j,:}\, , \mV_{i,j}\right)\begin{pmatrix}
                      1,\ldots, 1, -1
                  \end{pmatrix}^\top  \!.
    \end{aligned}
\end{equation}

\subsection{Factor Transformer Layer}
A Factor transformer layer (Factormer) is a message-passing layer that uses relation-aware multi-head attention with implicit edge features. The input is a graph with source node features $\vx_i\in \mathbb{R}^d$, target node features $x_j\in \mathbb{R}^d$, and edge features $\ve_{i,j}\in\mathbb{R}$. In an $\mH$-update based on the augmented line digraph from Section \ref{sec:graph_representation}, these correspond, in turn, to rows in $\mW$, rows in $\mH$, and elements in $\mV$. We denote this $\mH^{k+1}=\text{Factormer}(\mW^k, \mH^k, \mV)$, where $k$ is an iteration counter. 
For clarity of presentation, we omit biases and indices corresponding to the different heads.

First, we compute key, query, and value as in standard self-attention \citep{Vaswani2017}
\begin{equation}
    \vq_j = \vx_j\mTheta_\text{Q}, \quad \vk_i^\text{N} = \vx_i \mTheta_\text{K}^\text{N}, \quad \vv_i^\text{N} = \vx_i \mTheta_V^\text{N},
\end{equation}
where the weight matrices $\mTheta_\text{Q}, \mTheta_\text{K}^\text{N}, \mTheta_\text{V}^\text{N}$ are all square ($d\times d$). We will refer to the usual key and value as node key and node value, respectively, and indicate that using a superscript $\text{N}$. This is to distinguish those from the edge key and edge value, with superscript $\text{E}$, that we compute based on the implicit edge features defined in \eqref{eq:implicit_edge}
\begin{equation}
\vk_{i,j}^\text{E} = \tilde{\ve}_{i,j} \mTheta_\text{K}^\text{E}, \quad \vv_{i,j}^\text{E} = \tilde{\ve}_{i,j} \mTheta_V^\text{E},
\end{equation}
where, importantly, the weight matrices $\mTheta_\text{K}^\text{E}, \mTheta_\text{V}^\text{E}$ map to the same $d$-dimensional space as the node key and node value.

We compute the message $\vm_j$ by adding the node and edge values, multiplying with the attention weight, and finally performing sum aggregation
\begin{align}
    \alpha_{i,j} &= \text{softmax }\left(\vq_j\left( \mathbf{k}_i^\text{N} + \mathbf{k}_{i,j}^\text{E} \right )^\top   \middle/ \sqrt{d}\right),\\
    \vm_j &= \sum_{i\in\mathcal{N}(j)} \alpha_{i,j} \left ( \vv_i^\text{N} + \vv_{i,j}^\text{E}\right ).
\end{align}

The update function consists of adding the message to the node feature, applying layer norm $\text{LN}$ \citep{Ba2016}, passing it through a fully-connected feedforward network $\text{FFN}$, and finally adding it to the node feature and applying layer norm again, that is
\begin{equation}
    \vu(\vx_j) = \text{LN}\left(\vx_j+\text{FFN}\left(\text{LN}\left(\vx_j + \vm_j\right)\right)\right).
\end{equation}
 
\subsection{$N$-Factormer}
In the spirit of alternating minimization, we arrange multiple Factormers into a larger unit called an $N$-Factormer by first embedding the rows of $\mW$ and $\mH$ in a $d$-dimensional space, then composing $2N$ Factormer layers that alternate between updating each of the embedded factors while maintaining the same dimensionality, and finally mapping back to the original space 
\begin{equation}
    \begin{aligned}
        \tilde{\mH}^0 &= \mH^\top  \mTheta_\text{embed}, \quad \tilde{\mW}^0 = \mW^\top  \mTheta_\text{embed}, \\
        \tilde{\mH}^1 &= \text{Factormer}\left(\tilde{\mW}^0, \tilde{\mH}^0,  \mV\right), \\
        \tilde{\mW}^1 &= \text{Factormer}\left(\tilde{\mH}^1, \tilde{\mW}^0,  \mV^\top  \right), \\
        & \vdots \\
        \tilde{\mH}^N &= \text{Factormer}\left(\tilde{\mW}^{N-1}, \tilde{\mH}^{N-1},  \mV\right), \\
        \tilde{\mW}^N &= \text{Factormer}\left(\tilde{\mH}^N, \tilde{\mW}^{N-1},  \mV^\top  \right), \\
        \mH^{t+1} &= \tilde{\mH}^N\mTheta_\text{extract}, \quad \mW^{t+1} = \tilde{\mW}^N\mTheta_\text{extract}, 
    \end{aligned}
\end{equation}
where $\mTheta_\text{embed}\in\mathbb{R}^{r\times d}, \mTheta_\text{extract} \in\mathbb{R}^{d\times r}$ are weight matrices.

\subsection{Interleaved Optimization}
We evaluate two different acceleration models, ``learned intialization'' and ``learned acceleration''. Learned initialization only refines the initialization used to warm start ADMM, while learned acceleration is applied in between the ADMM updates. We summarize the key ideas below, further details---including pseudocode---can be found in the Appendix \ref{app:implementation}.

The learned initialization model, shown in Figure \ref{fig:overview_init_model}, maps the input $(\mW^0, \mH^0, \mV )$ through a $N$-Factormer and projects the output on the nonnegative orthant using a Rectified Linear Unit (ReLU). The outputs $(\mW^1, \mH^1)$ are then further improved using ADMM. 

The learned acceleration model, shown in Figure \ref{fig:overview_acc_model}, also uses a single $N$-Factormer and a ReLU, but arranged to perform neural fixed-point acceleration \citep{Venkataraman2021} for $T$ iterations. At each iteration, the inputs to the accelerator are (i) the previous output from the accelerator $(\mW^t,\mH^t)$, (ii) the previous output from ADMM $\left(\hat{\mW}^t,\hat{\mH}^t\right)$, and (iii) the edge feature matrix $\mV$. \citet{Venkataraman2021} suggested using a recurrent neural network for the acceleration, but we did not find any improved performance from that. 

\subsection{Loss}
Without additional assumptions, the NMF solution is not unique even after removing trivial symmetries \citep{Huang2013}. This makes supervised learning difficult. Instead we train in an unsupervised fashion, using a weighted sum of the objective function $\ell(\mW, \mH; \mV)$ over iterations $t=1,\ldots, T$ as the loss \citep{Andrychowicz2016}
\begin{equation}
    L\left(\{\mW^t,\mH^t\}_{t=1}^T, \mV\right) = \sum_{k=0}^{T} \gamma^k \,\ell(\mW^{T-k}, \mH^{T-k}, \mV),
\end{equation}
where $\gamma\in (0, 1]$ is a discount factor that places more weight on the loss at later iterations. Note that the final Euclidean projection step of our network guarantees that the output is always feasible.

\section{Experiments}
We implemented our models using Pytorch Geometric \citep{Fey2019}. The graph neural network is initialized using nonnegative SVD \citep{Boutsidis2008}, as implemented in NIMFA \citep{Zitnik2012}. We applied the acceleration for $T=5$ iterations and performed 5 ADMM iterations in each call. Both the learned initialization and the learned acceleration models used 4-Factormers with hidden dimension 100 and 4 self-attention heads. See Appendix \ref{app:implementation} for further details.

We will release code to reproduce the experiments if the paper is accepted.


\subsection{Data}

\subsubsection{Synthetic Data}

\begin{figure}
    \centering
    \includegraphics[width = 0.95\columnwidth]{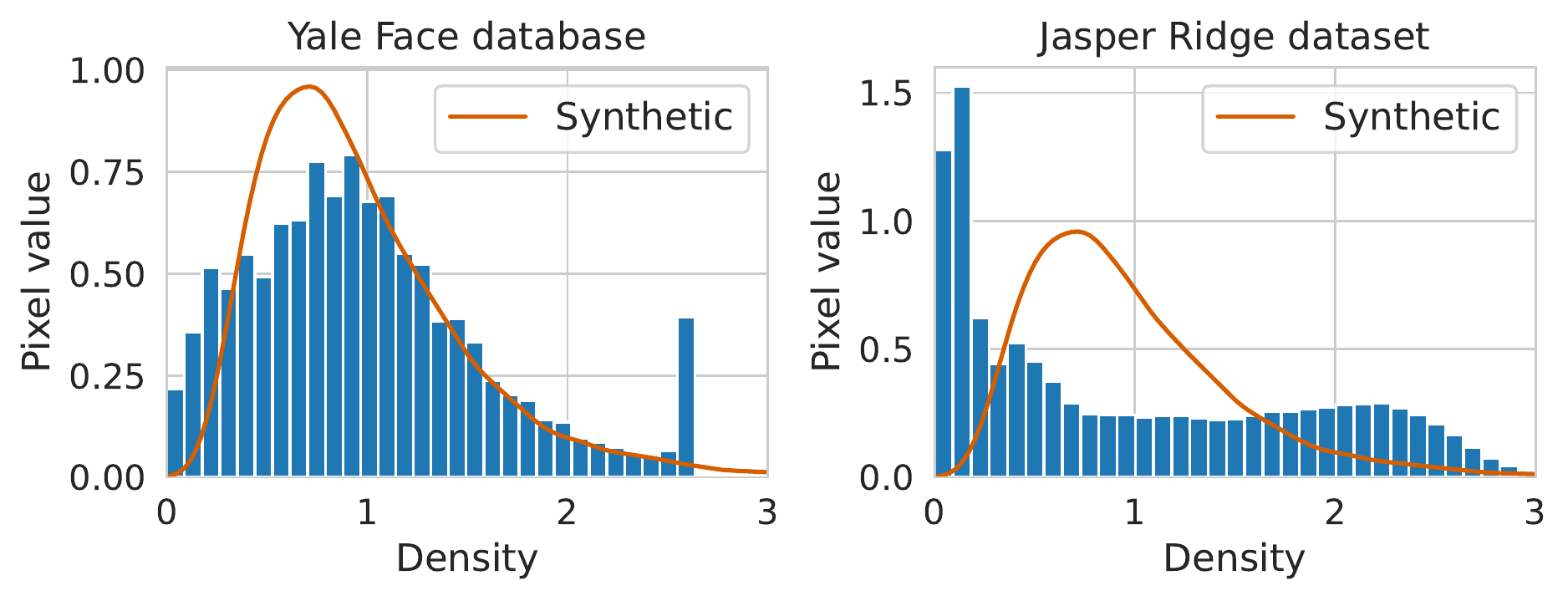}
    \caption{Histograms of pixel intensities for the Yale Face database and the Jasper Ridge dataset, together with the theoretical probability distribution of elements in the synthetic matrices.}
    \label{fig:data_dist}
\end{figure}
We train the model on synthetic data matrices $\mV$ defined as $\mV=\mW\mH^\top+\mN$, where the elements of $\mW$ and $\mH$ are sampled i.i.d. according to some predefined distribution and the noise is i.i.d. normally distributed, $\ermN_{i,j}\sim \mathcal{N}(0,\sigma^2)$. Obviously, we want the trained model to generalize to real-world data. Our initial (failed) experiments, using a Uniform distribution for the elements in $\mW$ and $\mH$, clearly showed that successful generalization requires a reasonable similarity between the values of the synthetic and real data matrices. We thus switched to sampling from an Exponential distribution with rate parameter $\lambda=1/\sqrt{r}$, chosen so to make the elements in the data matrix have mean 1, 
\begin{equation}
    \mathbb{E}\left[\ermV_{i,j}\right]=\mathbb{E}\left[\ermW_{i,:}\ermH_{j,:}^\top+\ermN_{i,j}\right]= r\cdot\lambda^2 + 0 = 1.
\end{equation}
This procedure is similar to the one of \citet{Huang2016} and, as shown in Figure \ref{fig:data_dist}, it roughly matches the distributions of the real data (normalized to have mean 1). 

We created a synthetic training dataset containing 15 000 matrices, out of which 10 000 were small, $n,m\sim \mathcal{U}[10,35]$, and the remaining 5 000 were larger, $n,m\sim \mathcal{U}[10,100]$ (see Appendix \ref{app:synthetic}). The underlying factors had rank $r=10$ but we also added noise with standard deviation $\sigma=0.01$. Additional details are provided in Appendix \ref{app:synthetic}. Similarly, we created two noise-free validation datasets: one with 128 small matrices, $n,m\sim \mathcal{U}[15,35]$, and one with 128 larger matrices, $n,m\sim \mathcal{U}[50,200]$. 

\subsubsection{Yale Face Database }
The Yale Face Database \cite{belhumeur1997eigenfaces} contains 165 grayscale images of size $64 \times 64$ pixels depicting 15 individuals. There are 11 images per subject, one per different facial expression or configuration (center-light, left-light, with glasses, without glasses, etc.). 

\subsubsection{Hyperspectral imaging}
The Jasper Ridge dataset \cite{rodarmel2002principal} is one of the most widely used datasets of hyperspectral images. We consider a sub-image of $100\times100$ pixels with 198 channels \citep{Zhu2017HyperspectralUG} (spectral bands), where some bands have been removed due to dense water vapor and atmospheric effects. 

\subsection{Convergence results}
Since the matrices have different sizes, we measure the average, element-wise, reconstruction error of the data matrices, which is equivalent to the root-mean-square error (RMSE). The RMSE can be interpreted, roughly, as the standard deviation of the prediction error. To appreciate the scale, recall that the data matrices have been normalized to have mean 1.

\begin{figure*}%
\centering
\begin{subfigure}{.499\columnwidth}
\includegraphics[width=\columnwidth]{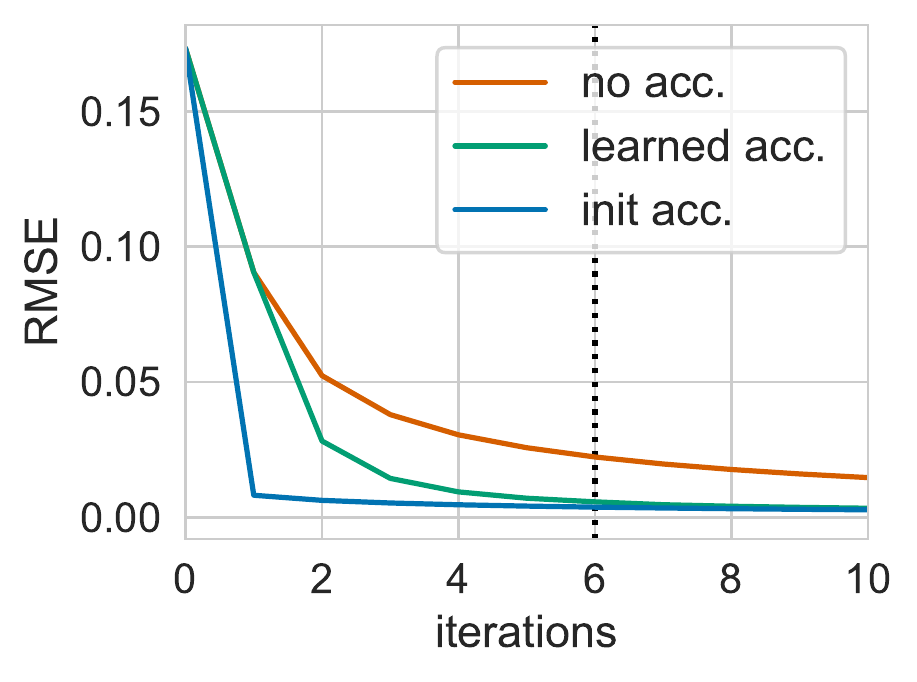}%
\caption{}%
\label{}%
\end{subfigure}%
\begin{subfigure}{.499\columnwidth}
\includegraphics[width=\columnwidth]{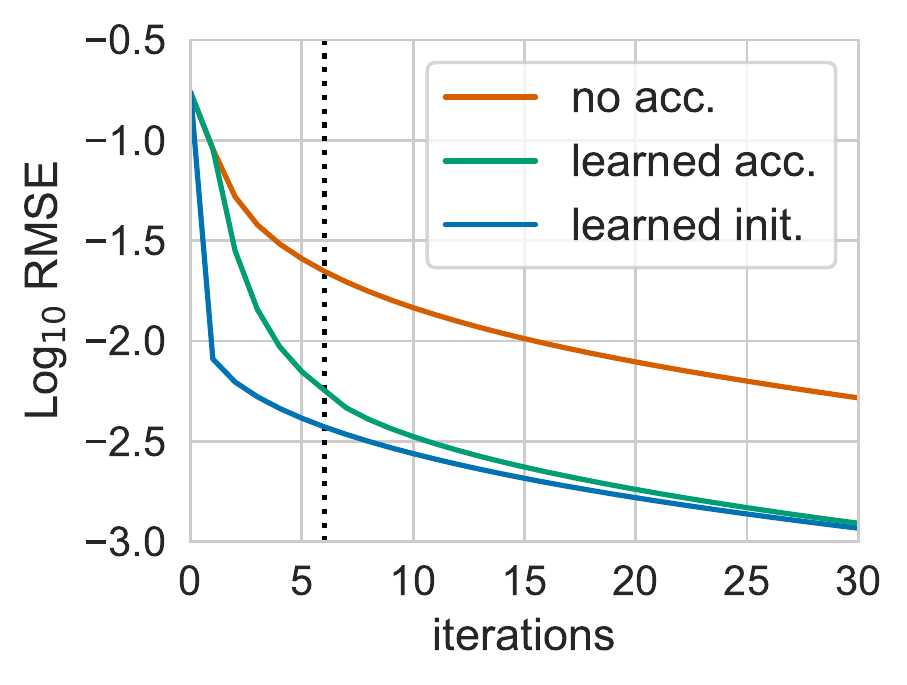}%
\caption{}%
\label{}%
\end{subfigure}%
\begin{subfigure}{.499\columnwidth}
\includegraphics[width=\columnwidth]{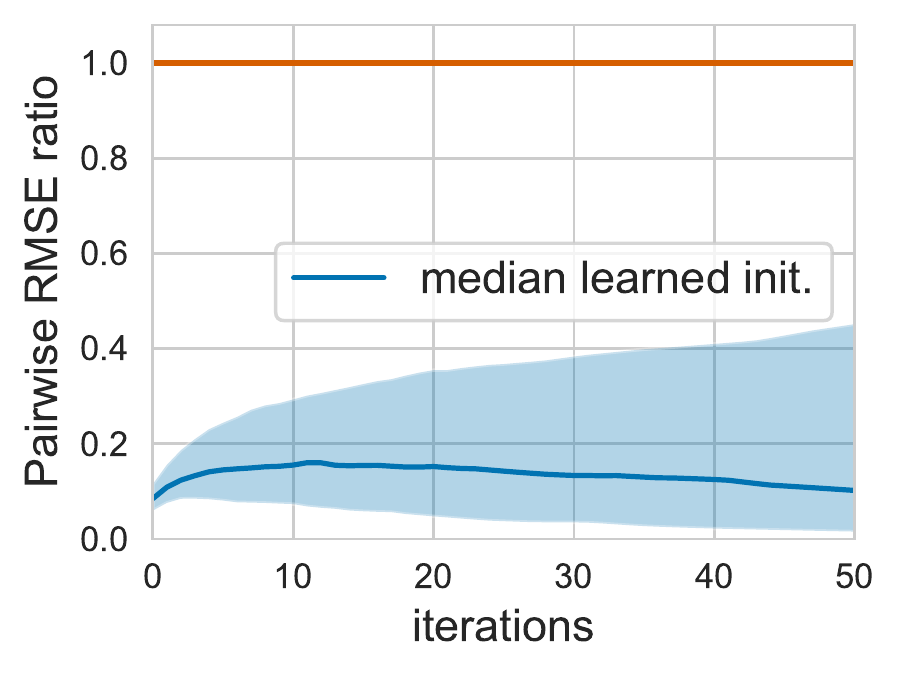}%
\caption{}%
\label{}%
\end{subfigure}%
\begin{subfigure}{.499\columnwidth}
\includegraphics[width=\columnwidth]{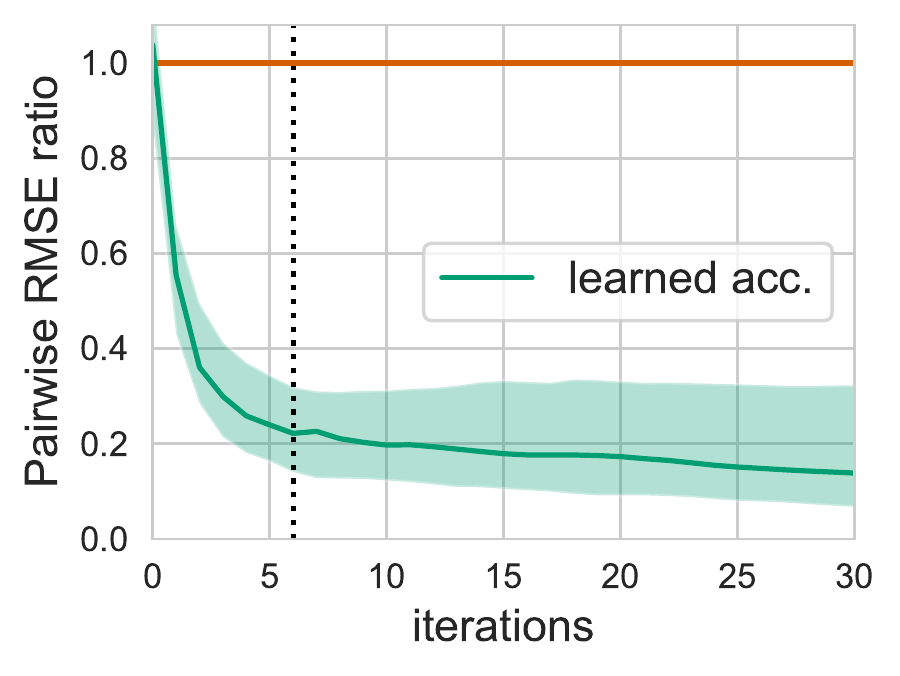}%
\caption{}%
\label{}%
\end{subfigure}
\caption{Results on the small synthetic validation dataset for the two accelerated methods and the baseline AO-ADMM methodl. a) The reconstruction error (RMSE) at each iteration; b) $\log_{10}$ RMSE; c):  the median and the quartiles of the pairwise ratio of reconstruction errors between the learned initialization model and the baseline; d):  the median and the quartiles of the pairwise ratio of reconstruction errors between the learned acceleration model and the baseline.}
\label{fig:synthetic_small}
\end{figure*}

\begin{figure*}%
\centering
\begin{subfigure}{.499\columnwidth}
\includegraphics[width=\columnwidth]{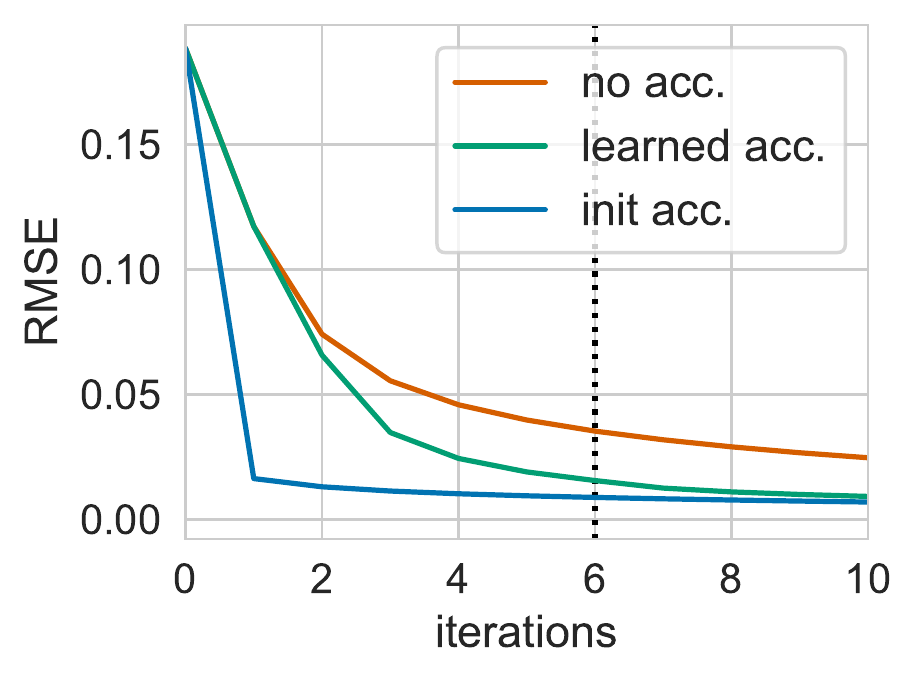}%
\caption{}%
\label{}%
\end{subfigure}%
\begin{subfigure}{.499\columnwidth}
\includegraphics[width=\columnwidth]{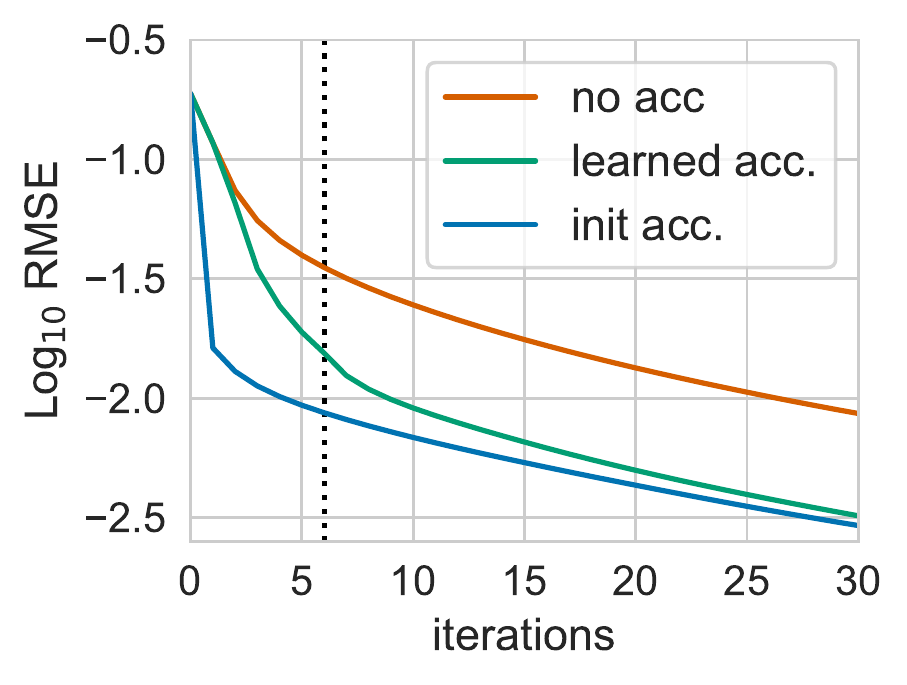}%
\caption{}%
\label{}%
\end{subfigure}%
\begin{subfigure}{.499\columnwidth}
\includegraphics[width=\columnwidth]{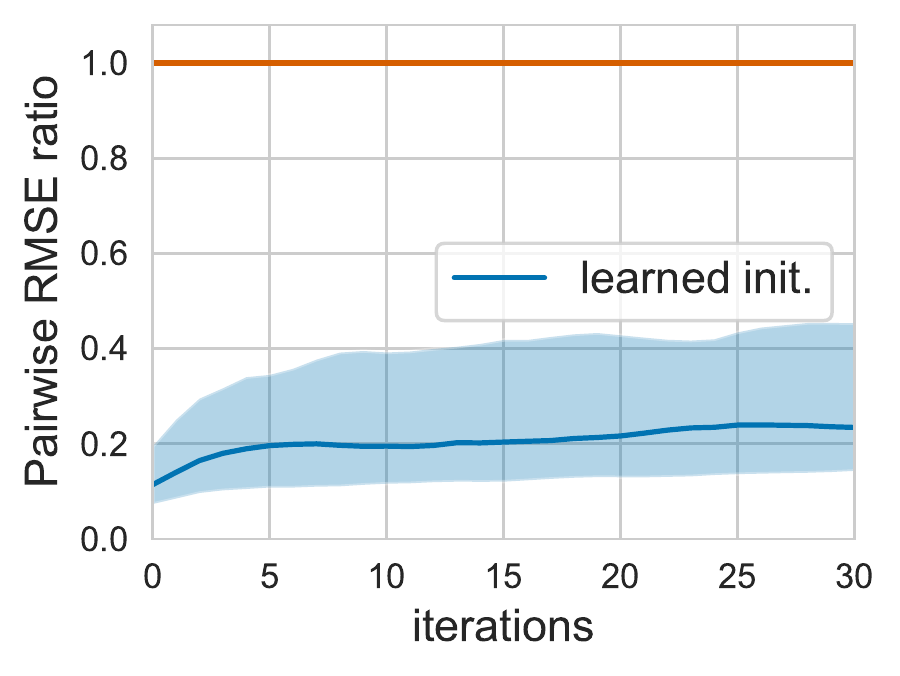}%
\caption{}%
\label{}%
\end{subfigure}
\begin{subfigure}{.499\columnwidth}
\includegraphics[width=\columnwidth]{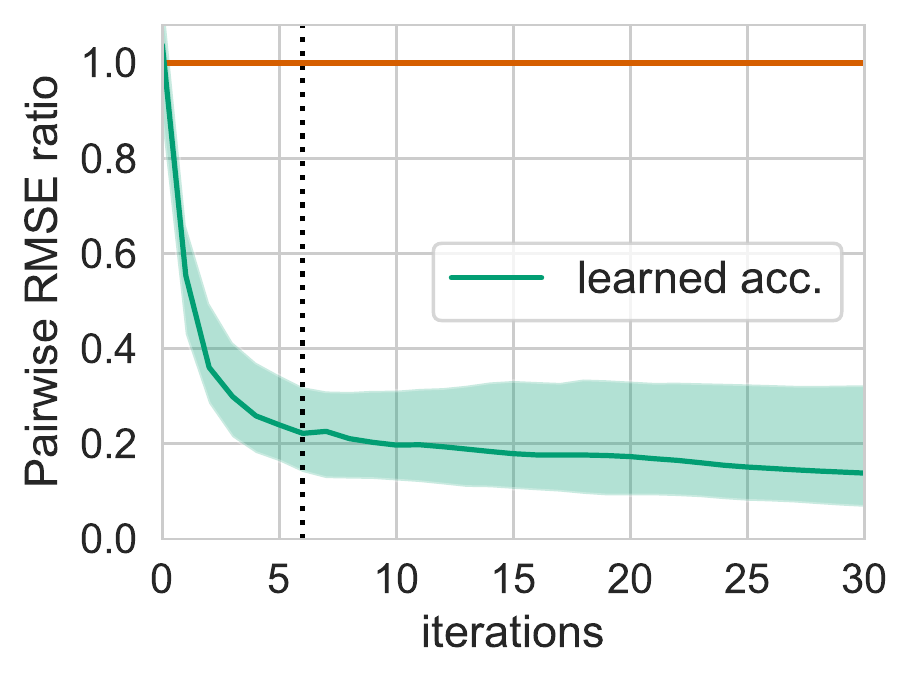}%
\caption{}%
\label{}%
\end{subfigure}%
\caption{Results on the large synthetic validation dataset for the two accelerated methods and the baseline AO-ADMM method. a) reconstruction error (RMSE) at each iteration; b) $\log_{10}$ RMSE; c): the median and the quartiles of the pairwise ratio of reconstruction errors between the learned initialization model and the baseline; d): the median and the quartiles of the  pairwise ratio of reconstruction errors between the learned acceleration model and the baseline.}
\label{fig:synthetic_large}
\end{figure*}

The results on the small and the large synthetic validation datasets are shown in Figures \ref{fig:synthetic_small} and \ref{fig:synthetic_large}, respectively. The two leftmost columns are plots of the RMSE at each iteration, on standard and semi-logarithmic scale. Each of the two rightmost plots compares the pairwise (same data matrix) RMSE ratio of an accelerated method to the one without acceleration. The solid line is the median (second quartile) and the shaded area corresponds to the results in between the first and third quartiles. The learned initialization performs notably well, immediately producing a fit that it takes the method without acceleration 20 iterations to match.

\begin{figure*}%
\centering
\begin{subfigure}{.5\columnwidth}
\vspace{0.15in}
\includegraphics[width=\columnwidth]{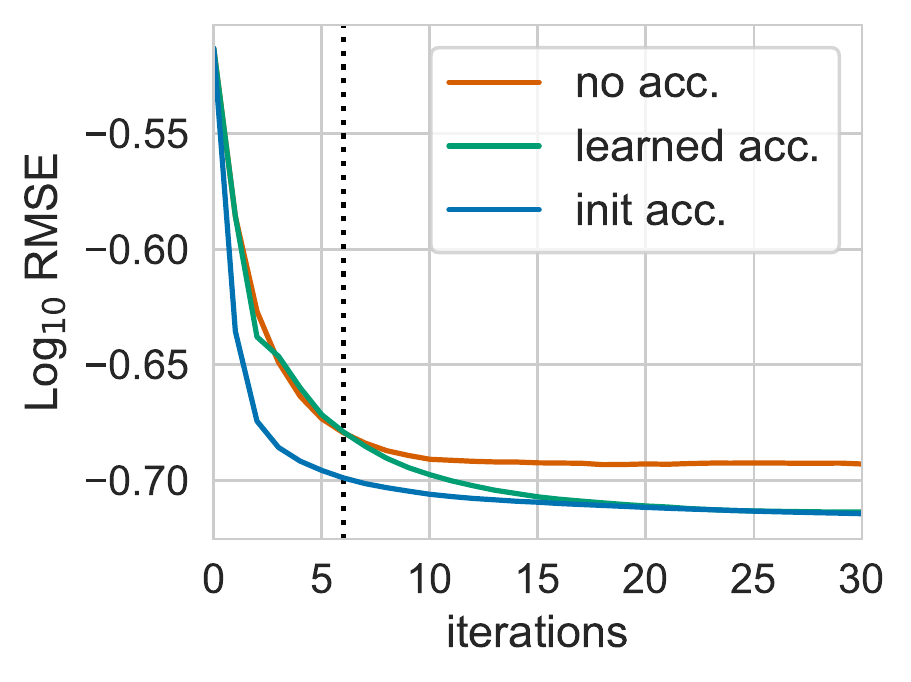}%
\caption{}%
\label{fig:faces_conv}%
\end{subfigure}%
\begin{subfigure}{.5\columnwidth}
\vspace{0.15in}
\includegraphics[width=\columnwidth]{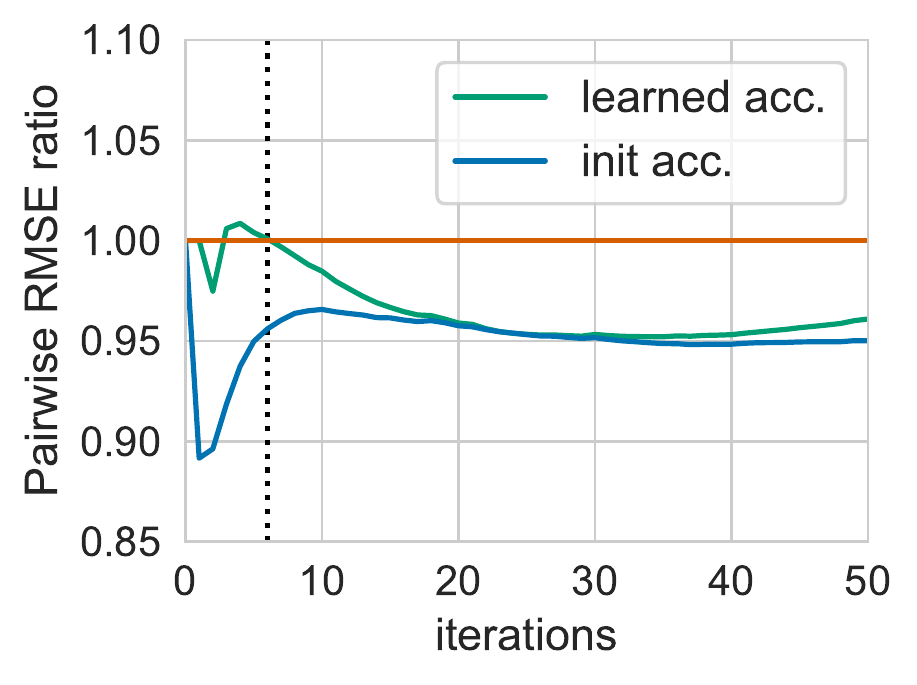}%
\caption{}%
\label{fig:faces_rel}%
\end{subfigure}\hspace{0.2in}%
\begin{subfigure}{.78\columnwidth}
\includegraphics[width=\columnwidth]{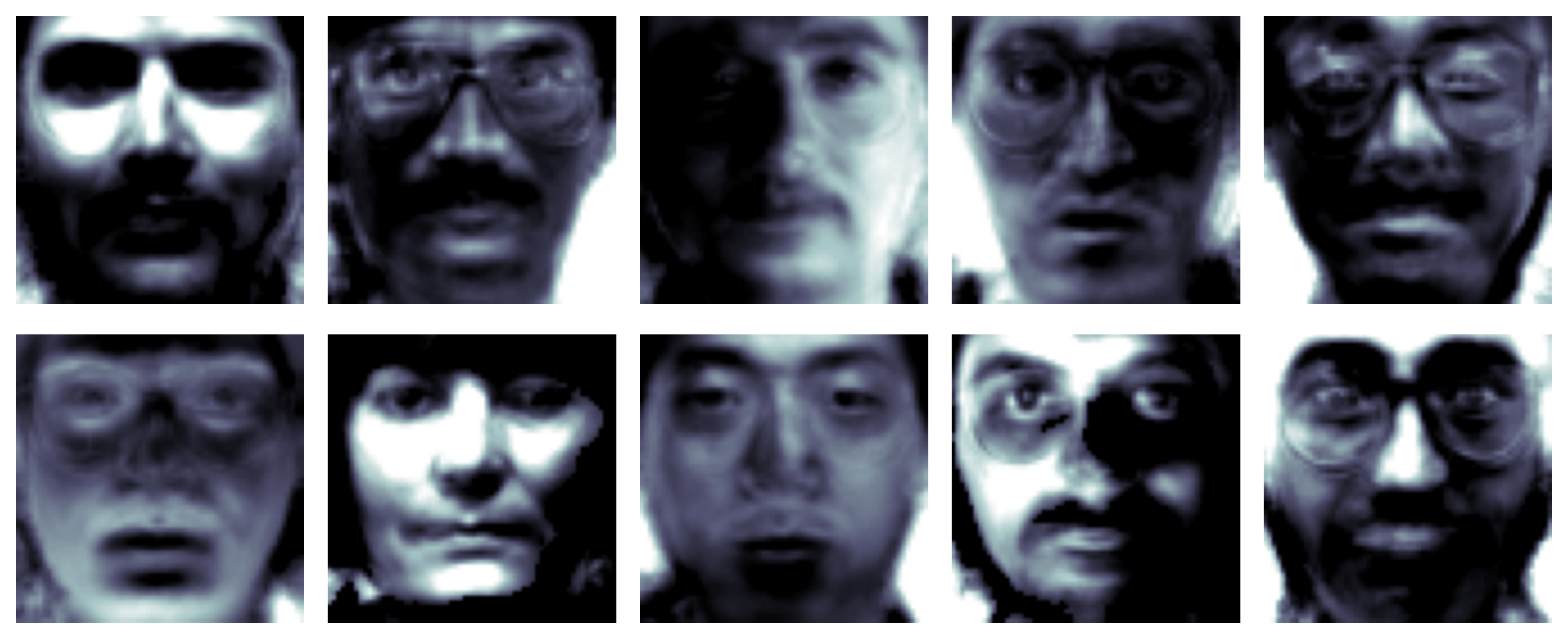}%
\caption{}%
\label{fig:faces_image}%
\end{subfigure}%

\caption{Yale Face Database. a): reconstruction errors (RMSE) at each iteration for the two accelerated methods and the baseline AO-ADMM method; b) : Pairwise  ratios of the reconstruction error of each accelerated method to the baseline; c): images corresponding to the ten basis vectors produced by the learned initialization model followed by 30 iterations of ADMM.}
\label{fig:faces}
\end{figure*}

\begin{figure*}%
\centering
\begin{subfigure}{.5\columnwidth}
\vspace{0.15in}
\includegraphics[width=\columnwidth]{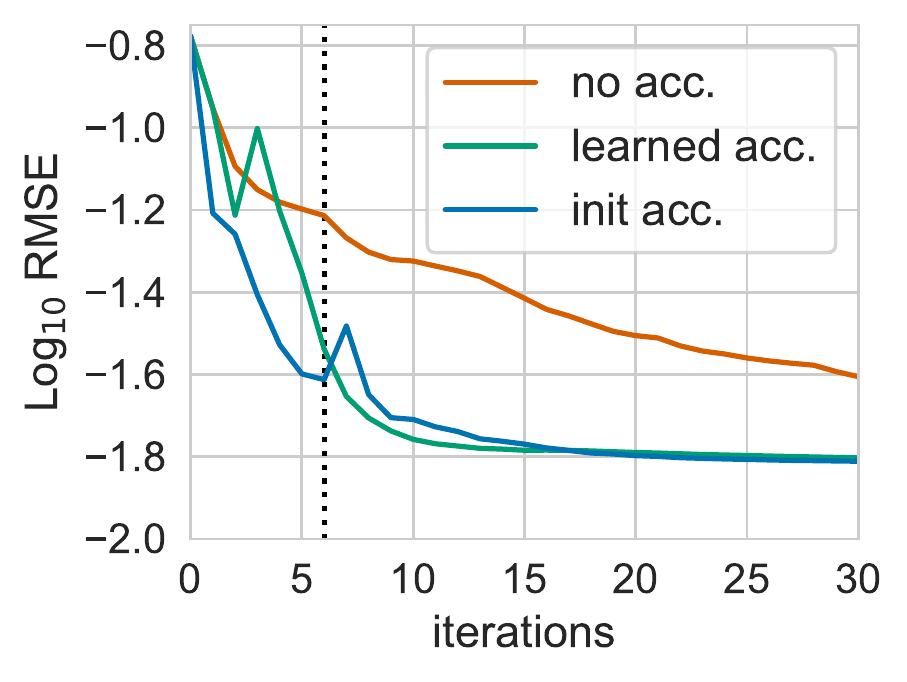}%
\caption{}%
\label{fig:hyperspectral_conv}%
\end{subfigure}%
\begin{subfigure}{.5\columnwidth}
\vspace{0.15in}
\includegraphics[width=\columnwidth]{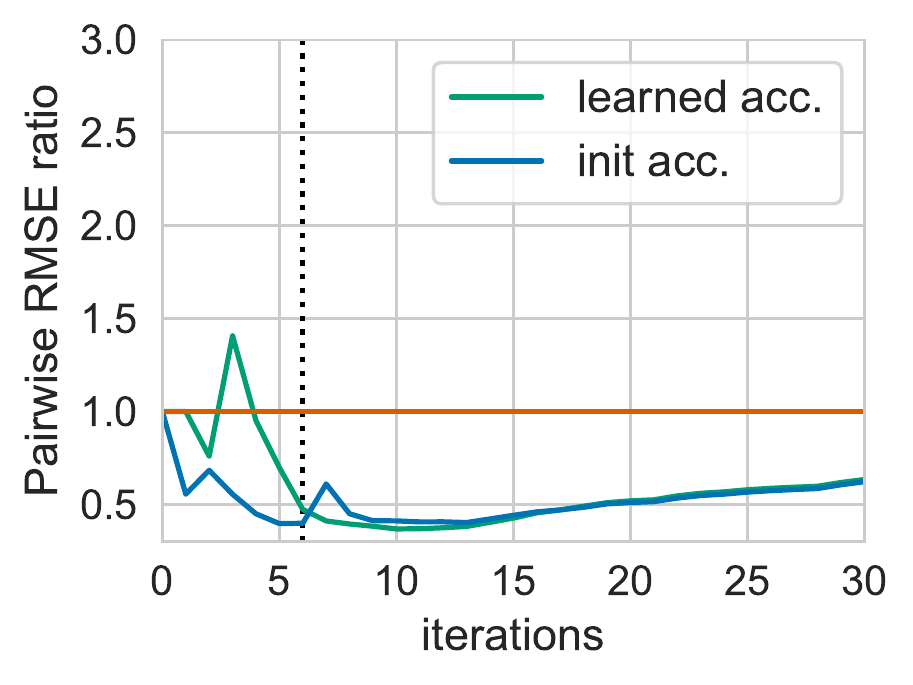}%
\caption{}%
\label{fig::hyperspectral_rel}%
\end{subfigure}\hspace{0.2in}%
\begin{subfigure}{.78\columnwidth}
\includegraphics[width=\columnwidth]{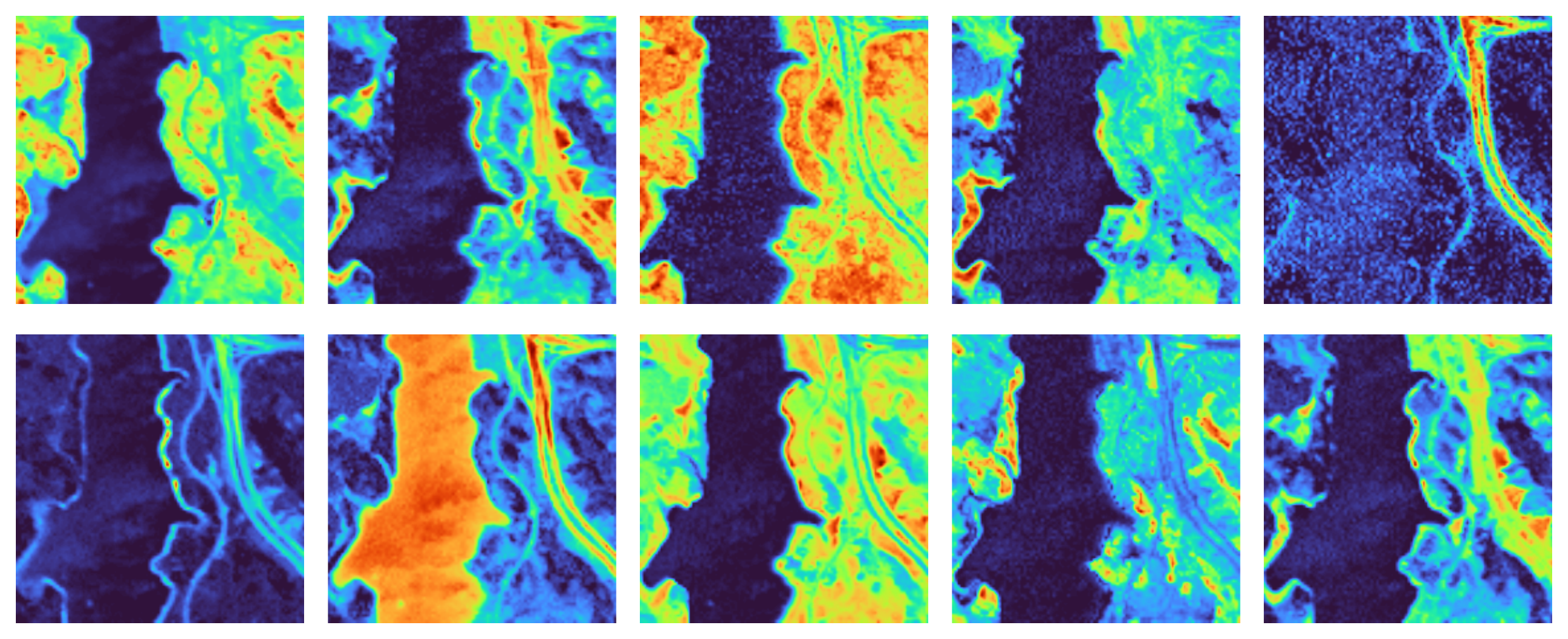}%
\caption{}%
\label{fig:spectral_image}%
\end{subfigure}%
\caption{Jasper ridge hyperspectral data set. a) Convergence plots for the two accelerated methods and the baseline AO-ADMM method; b) Pairwise ratios of the reconstruction error of each accelerated method to the baseline; c): Images corresponding to the ten basis vectors produced by the learned initialization model followed by 30 iterations of ADMM.}
\label{fig:hyperspectral}
\end{figure*}

Similarly, the results on the Yale Face database and the Jasper Ridge dataset are presented in Figures \ref{fig:faces} and \ref{fig:hyperspectral}. We also show the 10 basis vectors produced by the learned initialization followed by 30 ADMM iterations. On both of these real datasets, the accelerated methods seem to result in a better local optimum.

\section{Limitations}
There are two major limitations of this work that we are aware of. The first is that, since the rank corresponds directly to the dimensionality of the node features, the network can only be trained for a given rank. Potentially, this could be addressed by borrowing ideas used to create node embeddings \citep{Hamilton2017}. The second major limitation is computational scalability. Transformers in general are known to be computationally expensive, although remedies have been proposed \citep{Kitaev2020}. But another bottleneck is that the augmented line digraph is a \emph{complete} bipartite graph where the nodes either have degree $m$ or $n$, which makes the message-passing step computationally costly for large matrices. This is, however, a problem shared with many real-world graph problems \citep{Li2021}, which has therefore led to several promising strategies, including neighborhood sampling \citep{Hamilton2017,Chen2018} and graph sampling \citep{Zeng2019}. 




\section{Conclusions}
In this work, we have shown how the König digraph gives us a way of designing graph neural networks for numerical linear algebra and exemplified that by proposing an augmented line digraph suited to low-rank factorization problems. We further described a graph neural network architecture that interleaves Factormers with unrolled optimization steps and showed that it could substantially accelerate alternating optimization schemes for nonnegative matrix factorization.
We foresee that recent and ongoing research on graph neural networks will lead to even more capable architectures, as well as clever ideas to improve computational scalability, which would further extend the applicability of this work.


\bibliography{main}
\bibliographystyle{icml2021}

\newpage
\appendix
\onecolumn
\section{Implementation details}\label{app:implementation}
We present the pseudo-code of the learned initialization model and the learned acceleration model in Algorithm \ref{alg:init_net} and \ref{alg:acc_net}. We present the pseudo-code for the Factormer in Algorithm \ref{alg:factormer}. 

We used the AdamW optimizer \cite{loshchilov2017decoupled} for all experiments. We set the batch size to one. We trained the models using cosine annealing warm start \cite{loshchilov2016sgdr} where we let the learning rate decrease during the whole epoch. We set the initial learning rate to $10^{-4}$ for the initialization model and $10^{-5}$ for the learned acceleration model. After every epoch, we decreased the initial learning rate by a factor of 0.9. We trained both models during 15 epochs. For the learned acceleration model, we set the discount factor on the loss to $\gamma = 0.2$. 

When we trained the acceleration model, we first used one acceleration step. We then increased the number of acceleration steps by one at every two epochs until we reached five acceleration iterations. 
\begin{algorithm}[ht]
   \caption{Learned initialization model}
   \label{alg:init_net}
\begin{algorithmic}
   \STATE {\bfseries Input:} $\mathbf{W}^0$, $\mathbf{H}^0$, $\mV$, $T$ 
   \STATE $\mH^0 = \text{LinearEmbed}(\mH^0)$, $\mW^0 = \text{LinearEmbed}(\mW^0)$  \hfill\COMMENT{maps rank to hidden dimension.}
   \FOR{ Factormer in N-Factormer}
      \STATE $\mathbf{H}^0 = \text{Factormer}(\mW^0, \mH^0, \mathbf{V})$ 
 \STATE $\mathbf{W}^0 = \text{Factormer}(\mathbf{H}^0, \mathbf{W}^0, \mathbf{V}^T)$ 
   \ENDFOR
    \STATE $\mathbf{H}^1 = \text{LinearExtract}(\mathbf{H}^0)$ ,  $\ \mathbf{W}^1 = \text{LinearExtract}(\mathbf{W}^0)$ \hfill\COMMENT{maps hidden dimension to rank.}
    \FOR{$t = 1\hdots T$} 
    \STATE ${\mH}^{t+1} = \text{Solver}(\mW^t, \mH^t, \mV)$
   \STATE ${\mW}^{t+1} = \text{Solver}(\mH^{t+1}, \mW^t, \mV^T)$
   \ENDFOR
\end{algorithmic}
\end{algorithm}
\begin{algorithm}[ht]
   \caption{Learned acceleration model}
   \label{alg:acc_net}
\begin{algorithmic}
   \STATE {\bfseries Input:} $\mathbf{W}^0$, $\mathbf{H}^0$, $\mV$, $T$, nbrAcc \hfill\COMMENT{nbrAcc = number of acceleration steps}
   \FOR{$t = 0\hdots T-1$} 
    \STATE $\hat{\mH}^t = \text{Solver}(\mW^t, \mH^t, \mV)$
   \STATE $\hat{\mW}^t = \text{Solver}(\hat{\mH}^t, \mW^t, \mV^T)$
   \IF{$t< \text{nbrAcc}$}
   \STATE $\mathbf{H}^t = \text{Concatenate}(\mathbf{H}^t, \widehat{\mathbf{H}}^t)$, $ \ \mathbf{W}^t = \text{Concatenate}(\mathbf{W}^t, \hat{\mathbf{W}}^t)$
   \STATE $\mathbf{H}^t = \text{LinearEmbed}(\mathbf{H}^t)$ , $\ \mathbf{W}^t = \text{LinearEmbed}(\mathbf{W}^t)$ \hfill\COMMENT{maps rank$\times 2$ to hidden dimension.}
   \FOR{ Factormer in N-Factormer}
   \STATE $\mathbf{H}^t = \text{Factormer}(\mathbf{W}^t, \mathbf{H}^t,\mathbf{V})$ 
   \STATE $\mathbf{W}^t = \text{Factormer}(\mathbf{H}^t, \mathbf{W}^t, \mathbf{V}^T)$ 
   \ENDFOR
    \STATE $\mathbf{H}^{t+1} = \text{LinearExtract}(\mathbf{H}^t)$, $\ \mathbf{W}^{t+1} = \text{LinearExtract}(\mathbf{W}^t)$   \hfill\COMMENT{maps hidden dimension to rank.}
    \ELSE
    \STATE $\mathbf{H}^{t+1} = \hat{\mH}^t$
    \STATE $\mathbf{W}^{t+1} = \hat{\mW}^t$
    \ENDIF
    \ENDFOR
\end{algorithmic}
\end{algorithm}

\begin{algorithm}[ht]
   \caption{Factormer}
   \label{alg:factormer}
\begin{algorithmic}
   \STATE {\bfseries Input:} $\mathbf{x}_{i}$, $\mathbf{x}_{j}$, $\mathbf{e}_{i,j}$ \hfill\COMMENT{Source nodes, target nodes and edge attributes.}
   \STATE{\bfseries Derive message:}
   \STATE $\tilde{\mathbf{e}}_{i,j} = \left[\vx_i \odot \vx_j, \ve_{i,j}\right]$
   \STATE $\vq_j = \text{LinearQuery}(\vx_j)$, $\vk_i^N = \text{LinearNodeKey}(\vx_i)$, $\vk_i^E = \text{LinearEdgeKey}(\ve_{i,j})$
   \STATE  $\vv_i^N = \text{LinearNodeValue}(\vx_i)$, $\vv_i^E=\text{LinearEdgeValue}(\ve_{i,j})$
   \STATE $\alpha_{i,j} = \text{softmax }\left(\vq_j\left( \mathbf{k}_i^\text{N} + \mathbf{k}_{i,j}^\text{E} \right )^\top   \middle/ \sqrt{d}\right)$ \hfill\COMMENT{Derive self attention.}
   \STATE $\mathbf{m}_{i,j} = \alpha_{i,j} \left ( \vv_i^\text{N} + \vv_{i,j}^\text{E}\right )$\hfill\COMMENT{Derive message.}
    \STATE{\bfseries Aggregate:}
    \STATE $\vx_j' =\sum_{i\in\mathcal{N}(j)} \vm_{i,j} $ \hfill\COMMENT{Add aggregation.}
   \STATE{\bfseries Update:}
   \STATE $\mathbf{x}_j =\mathbf{x}_j' + \mathbf{x}_j$
   \STATE $\mathbf{x}_j = \text{LayerNorm}(\mathbf{x}_j)$
    \STATE $\mathbf{x}_j =\text{FeedForward}(\mathbf{x}_j)$
   \STATE $\mathbf{x}_j =\mathbf{x}_j' + \mathbf{x}_j$
    \IF{ not last Factormer}
    \STATE $\mathbf{x}_j = \text{LayerNorm}(\mathbf{x}_j)$\hfill\COMMENT{In the last N-Factormer layer we skip the last layer norm.}
    \ENDIF
\end{algorithmic}
\end{algorithm}

\section{ADMM for nonnegative least-squares}\label{sec:ADMM}
To derive the alternating direction method of multipliers (ADMM) equations for the nonnegative least-squares problem in Equation \ref{eq:subproblem}, we introduce an auxiliary variable $\tilde{\mH}\in\mathbb{R}^{n\times r}$ and rewrite the problem as
\begin{equation}
    \begin{aligned}
        & \underset{\mH}{\text{minimize}}
        & & \frac{1}{2}\|\mW\mH^\top   - \mV\|_\text{F}^2 + I_{\tilde{\mH}\geq 0}(\tilde{\mH}) \\
        & \text{subject to} & & \mH - \tilde{\mH} = 0,
    \end{aligned}
\end{equation}
where $I_{\tilde{\mH}\geq 0}$ is an indicator function which is zero when $\tilde{\mH}\geq 0$ and infinite otherwise. Using an augmented Lagrangian parameter $\rho> 0$ (we use $\rho=1$) and a scaled dual variable $\mU\in\mathbb{R}^{n\times r}$, an iteration of ADMM then consists of the following equations,
\begin{align}
    \mH&\gets \left(\mW^\top\mW + \rho I\right)^{-1}\left(\mW^\top \mV + \rho (\tilde{\mH} + \mU) \right),\\
    \tilde{\mH} &\gets \left(\mH - \mU\right)_+,\\
    \mU &\gets \mU + \mH-\tilde{\mH},
\end{align}
where $(\cdot)_+$ denotes the nonnegative part of the expression, i.e. a projection on the nonnegative orthant. For computational efficiency, the factorization of $\mW^\top\mW + \rho I $ can be cached.

\section{Synthetic dataset}\label{app:synthetic}
We created the training data by concatenating small and large matrices. In Table \ref{tab:training_data} we describe in detail which matrices as included in the training data.
\begin{table}[t]
\caption{The sizes of the matrices in the training dataset. The total number of samples in the training dataset is 15000. The number or rows and columns in the different matrices are sampled uniformly as $n,m\sim \mathcal{U}[low,high]$.}
\label{tab:training_data}
\vskip 0.15in
\begin{center}
\begin{small}
\begin{sc}
\begin{tabular}{lcccr}
\toprule
Nbr of samples & Row range & Column range\\
\midrule
10000& $[10, 35]$& $(10, 35)$ \\
2000&[30, 70]&[30, 70]\\
1500&[30,100]&[10,35] \\
1500&[10,35]&[30,100]\\
\bottomrule
\end{tabular}
\end{sc}
\end{small}
\end{center}
\vskip -0.1in
\end{table}

\section{Additional results}
The root-mean square error (RMSE) is related to the Frobenius norm as follows
\begin{equation}
    \text{RMSE} = \|\mW\mH^\top - \mV\|_\text{F}/\sqrt{mn}.
\end{equation}

\begin{figure}[ht]
\centering
\begin{subfigure}{.9\linewidth}
\centering
\includegraphics[width=0.7\linewidth]{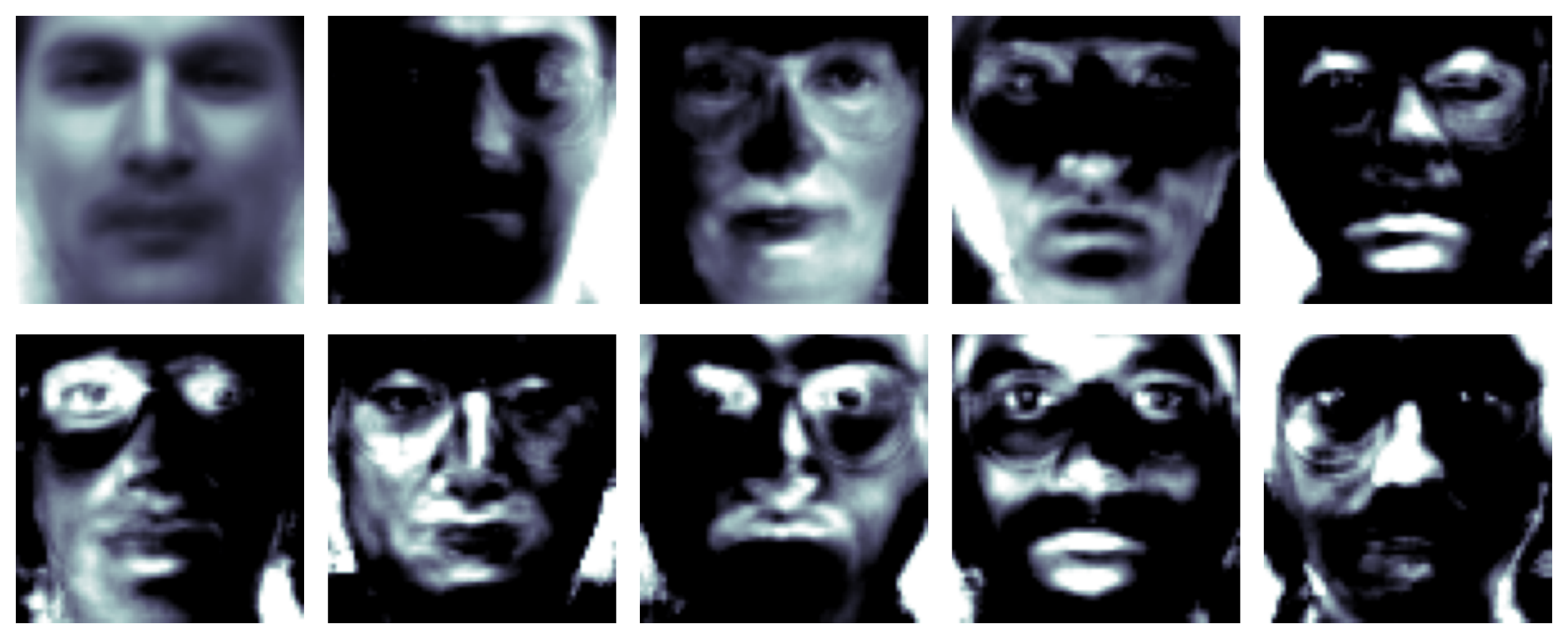}%
\caption{Only nonnegative SVD.}%
\end{subfigure}\hfil
\vspace{0.2in}
\begin{subfigure}{.9\linewidth}
\centering
\includegraphics[width=0.7\linewidth]{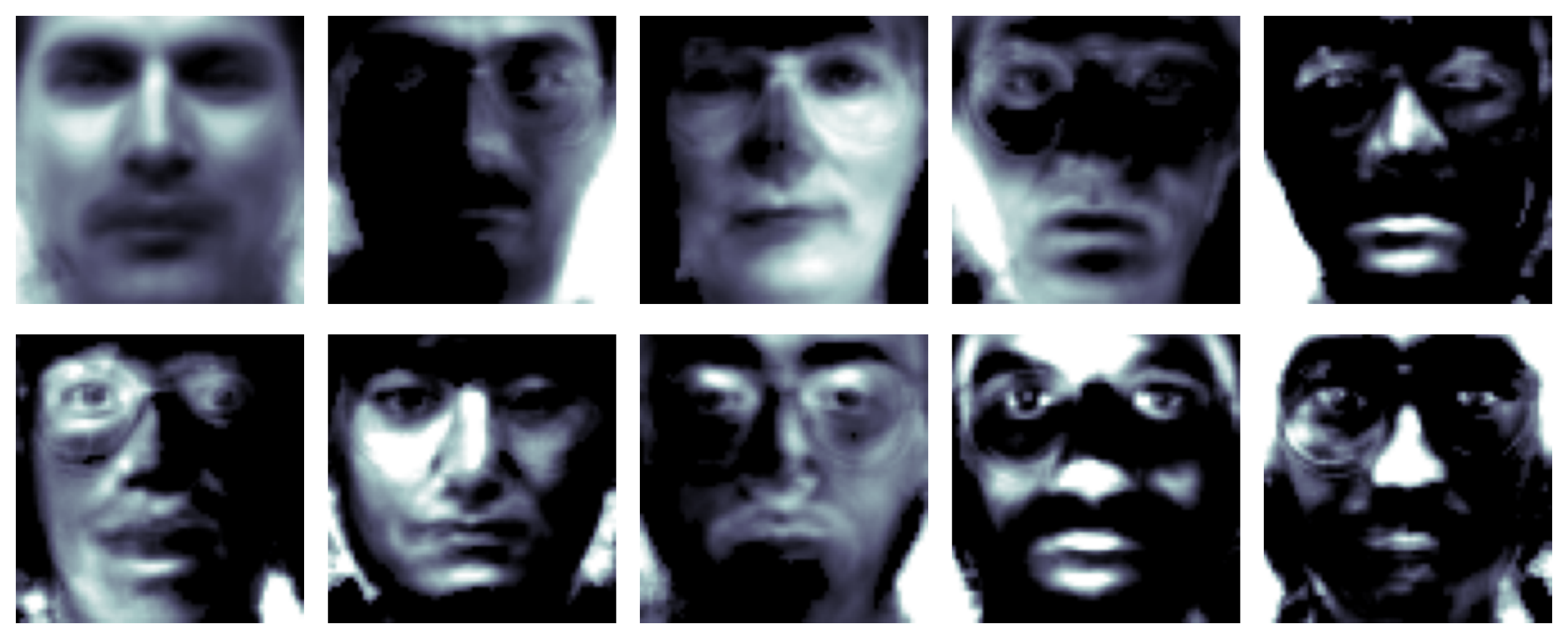}%
\caption{Initialization network followed by two ADMM iteration.}%
\end{subfigure}\hfil
\vspace{0.2in}
\begin{subfigure}{.9\linewidth}
\centering
\includegraphics[width=0.7\linewidth]{fig/faces_images.pdf}%
\caption{Initialization network followed by 30 ADMM iterations.}%
\end{subfigure}%
\caption{Images corresponding
to the ten basis vectors produced by: only nonnegateve SVD (a), the learned initialization model followed by two ADMM iterations (b), the learned initialization model followed by 30 iterations of ADMM.}
\label{fig:many_faces}
\end{figure}

\end{document}